\documentclass[twoside]{article}
\usepackage[accepted]{aistats2025}

\usepackage[round]{natbib}

\usepackage{amsfonts}
\usepackage{booktabs}
\usepackage{multirow}
\usepackage{graphicx}
\usepackage[colorlinks=true, linkcolor=black, citecolor=black, urlcolor=black]{hyperref}

\newtheorem{definition}{Definition}[section]

\begin{document}

\twocolumn[
\aistatstitle{TempTest: Local Normalization Distortion and the Detection of Machine-generated Text}
\aistatsauthor{Tom Kempton\textsuperscript{\ensuremath{\star}}\And Stuart Burrell\textsuperscript{\ensuremath{\star}} \And  Connor Cheverall\textsuperscript{\ensuremath{\dagger}}}
\aistatsaddress{Department of Mathematics\\ University of Manchester \And  Innovation Lab \\ Featurespace \And Institute of Astronomy\\ University of Cambridge}
]

% For additional author information (equal contribution, internship context), we have followed the exact latex code (https://arxiv.org/src/2311.07565) of https://proceedings.mlr.press/v238/janz24a/janz24a.pdf which appeared in PMLR for AISTATS 2024. 

\begin{abstract}
Existing methods for the zero-shot detection of machine-generated text are dominated by three statistical quantities: log-likelihood, log-rank, and entropy. As language models mimic the distribution of human text ever closer, this will limit our ability to build effective detection algorithms. To combat this, we introduce a method for detecting machine-generated text that is entirely agnostic of the generating language model. This is achieved by targeting a defect in the way that decoding strategies, such as temperature or top-k sampling, normalize conditional probability measures. This method can be rigorously theoretically justified, is easily explainable, and is conceptually distinct from existing methods for detecting machine-generated text. We evaluate our detector in the white and black box settings across various language models, datasets, and passage lengths. We also study the effect of paraphrasing attacks on our detector and the extent to which it is biased against non-native speakers. In each of these settings, the performance of our
test is at least comparable to that of other state-of-the-art text detectors, and in some cases, we strongly outperform these baselines.  
\end{abstract}
\section{INTRODUCTION}

There has been a surge of interest in statistical methods for the detection of machine-generated text since the paradigm shift in language modeling triggered by the transformer architecture \citep{vaswani2017attention}. This task has growing societal importance, for example, in combating fake news \citep{ahmed2021detectingfakenewsusing} or ensuring academic integrity \citep{education}.

\begin{NoHyper}
	\renewcommand{\thefootnote}{\ensuremath{\star}}
	\footnotetext{Equal contribution.}
        \renewcommand{\thefootnote}{\ensuremath{\dagger}}
	\footnotetext{Work completed while an intern at Featurespace.}
\end{NoHyper}

Approaches based on supervised learning typically struggle with out-of-domain generalization \citep{baofast}, leading research to focus on the \emph{zero-shot} regime. To date, state-of-the-art zero-shot techniques are typically based on three fundamental statistical quantities: log-likelihood, log-rank, and entropy \citep{gehrmann2019gltr}. Although state-of-the-art methods such as DetectGPT \citep{mitchell2023detectgpt}, Fast-DetectGPT \citep{baofast}, DNA-GPT \citep{yangdna}, and Detect-LLM \citep{su2023detectllm} have made remarkable progress using these measures, our work demonstrates that a wider suite of basic statistical quantities is beneficial.

We introduce a new zero-shot detection method, TempTest, based on a novel statistical quantity, TempNorm. This statistic is derived from the distortion of conditional probability measures caused by the way that temperature sampling renormalizes probability mass. Temperature sampling is a ubiquitous decoding strategy that allows fine control over the diversity or creativity of generated text. Similar techniques extend to top-k and nucleus (top-p) sampling, although the signal for top-p is very small; see Appendices \ref{app:top-k} and \ref{app:topp}. 

By explicitly tying detection to artifacts of the text generation process, we are agnostic to the large language model used for generation. This may be important for modern large language models such as Llama 3.1 \citep{dubey2024llama}, which in Section \ref{sec:experiments} we demonstrate cause challenges to existing detectors but not TempTest. In addition, TempTest is easily explainable and rigorously mathematically justified, which are important features for societal adoption given the increasing regulation around the deployment and monitoring of AI models, such as the EU AI act.

In summary, the contributions of this paper are to:
\begin{itemize}
    \item Introduce TempTest, a new state-of-the-art zero-shot machine-generated text detection method, based on quantifying local normalization distortion of conditional probability measures.
    \item Validate the robustness of this method across a range of datasets, language models, hyperparameter choices, and white box and black box settings.
    \item Derive a rigorous justification of this technique based on Bayesian statistics and ergodic theory, showing that it is intuitive and easily explainable.
\end{itemize}
We emphasize that while TempTest achieves comparable or superior performance to existing art, we do not see ourselves in competition with other detection methods. Instead, our hope is that broadening the set of tools available to our community with a genuinely new approach opens the possibility for nuanced combinations and even better results. 

\section{BACKGROUND}\label{sec:background}

%%%%%%%%%%%%%%%%%%%%%%%%%%%%%%%%%%%%%%%%%%%%%%%%%%%
\paragraph{Problem Formulation}
%%%%%%%%%%%%%%%%%%%%%%%%%%%%%%%%%%%%%%%%%%%%%%%%%%%

We treat the problem of detection of machine-generated text as a binary classification problem. Given a passage of text ${\underline w}= w_1\dots w_T$ of length $T \in \mathbb{N}$, one wishes to assign labels of $0$ and $1$ for human and machine-generated text, respectively, via a classifier $f({\underline w})$. We work in a zero-shot setting, where $f$ does not have learnable parameters optimized via a training process. 

%%%%%%%%%%%%%%%%%%%%%%%%%%%%%%%%%%%%%%%%%%%%%%%%%%%
\paragraph{Decoding Strategies}
%%%%%%%%%%%%%%%%%%%%%%%%%%%%%%%%%%%%%%%%%%%%%%%%%%%

Given an auto-regressive language model $P$ with vocabulary $\mathcal V$ of size $|\mathcal V|=N$ and context $\underline{w} = w_{i-L}\cdots w_{i-1} \in {\mathcal V}^L$, 
$$
\textnormal{Softmax}(\text{Logits}(\underline{w})) = \exp(l_i) / {\sum\limits_{j=1}^{N} \exp(l_j)}
$$
transforms the output Logits$(\underline{w}) = (l_1,\dots,l_N)$ into a probability measure $p(\cdot|w_{i-L}\cdots w_{i-1})$ over the set of predicted next tokens from $\mathcal V$. We refer to the process of sampling directly from the distributions $p$ as \emph{pure-sampling}. For concision we denote $w_{<i}:=w_{i-L}\cdots w_{i-1}$.

In practice, during inference, different decoding strategies are used to give finer control over the qualitative nature of the output, for example by affecting the trade off between quality and diversity \citep{caccia2019, zhang2021trading}. For example, \emph{top-k} sampling \citep{fan2018hierarchical} samples only over a renormalized truncated distribution of the top $k$ tokens with the highest conditional probability. More precisely, given a context $w_{<i}$ and a value of $k>1$, the associated top-k set $\mathcal V_k(w_{<i})$ is the subset of $\mathcal V$ consisting of the $k$ tokens assigned highest probability by the measure $p(\cdot|w_{<i})$. Top-k sampling assigns mass zero to all tokens outside of the top-k set, while giving tokens inside the top-k set mass
\begin{equation}\label{topksampling}
q_k(v_i|w_{<i}):=\dfrac{p(v_i|w_{<i})}{\sum_{v\in\mathcal V_k(w_{<i})}p(v|w_{<i})}.
\end{equation}
Top-p sampling \citep{Holtzman2020The} may be defined analogously, and we omit for brevity.

Our particular focus is on text generated through temperature sampling \citep{guo2017calibration}. Given a temperature parameter $\tau\in(0,1]$, the probability distribution $Q_{\tau}$ resulting from sampling from language model $P$ at temperature $\tau$ is given by
\begin{equation}\label{tempsampling}
q_{\tau}(v_i|w_{<i}):=\dfrac{(p(v_i|w_{<i}))^{\frac{1}{\tau}}}{\sum_{v\in\mathcal V}(p(v|w_{<i})^{\frac{1}{\tau}}}.
\end{equation}

Temperature, top-k and nucleus sampling are widely used as methods of reducing the chance of sampling from the tail of the probability distribution $p(\cdot|w_{<i})$ \citep{meister2023efficacy}. While helpful, these decoding strategies may result in statistically significant traces due to distortions in the renormalized conditional distributions that detection algorithms may exploit.

%%%%%%%%%%%%%%%%%%%%%%%%%%%%%%%%%%%%%%%%%%%%%%%%%%%
\paragraph{Quantifying Statistical Properties Of Text}
%%%%%%%%%%%%%%%%%%%%%%%%%%%%%%%%%%%%%%%%%%%%%%%%%%%

Recall that, given a sequence of tokens $w_1\cdots w_T$ that we wish to classify, most existing zero-shot detectors are based on log-likelihood, log-rank, or entropy. We give mathematical definitions of these quantities here to highlight how they compare and contrast to our new quantity, TempNorm, introduced in Section \ref{sec:TempTest}.

The \emph{per-token log-likelihood} of the sequence $w_1\cdots w_T$ is given by
\[
\frac{1}{T}\log P(w_1\cdots w_T)=\frac{1}{T}\log\left(\prod_{i=1}^Tp(w_i|w_{<i})\right).
\]
 
Note that the often used metric \emph{perplexity} is the exponential of the negative per-token log-likelihood. The {\it rank} $r(v_i|w_{<i})$ of the token $v_i$ given context $w_{<i}$ is the position of token $v_i$ in the list of tokens ordered by decreasing probability $p(v|w_{<i})$, with the most likely token having rank $1$. Analogously, the {\it per-token log-rank} of $w_1\cdots w_T$ is given by
\[
\frac{1}{T}\log r(w_1\cdots w_T)=\frac{1}{T}\log\left(\prod_{i=1}^T r(w_i|w_{<i})\right).
\]

Finally, the \emph{per-token entropy} is given by
\[
\frac{1}{T}H(w_1\cdots w_T)=\frac{1}{T}\sum_{i=1}^T\left(\sum_{v\in\mathcal V} p(v|w_{<i})\log p(v|_{w_{<i}})\right).
\]

%%%%%%%%%%%%%%%%%%%%%%%%%%%%%%%%%%%%%%%%%%%%%%%%%%%
\section{RELATED WORK}\label{sec:related_work} 
%%%%%%%%%%%%%%%%%%%%%%%%%%%%%%%%%%%%%%%%%%%%%%%%%%%

There are two main approaches to the detection of machine-generated text. The first is to train a supervised classifier, which are often excellent in domain but their performance can degrade heavily when the language model or text domain changes \citep{baofast}. Some recent works have begun to address this challenge \citep{li-etal-2024-spotting,zhang2024detecting}, though still face the limitation of requiring training processes and data. The second, which we study in this article, is the zero-shot approach. State-of-the-art zero-shot approaches are discussed below.
We also mention watermarking, which is the practice of modifying the output of a language model in order to leave specific statistical traces which are easy to detect, see, for example, \cite{kirchenbauer2023watermark}. Watermarking works well, but only when incorporated directly into the language model. For a survey of efforts to detect machine-generated text see \cite{ghosal2023survey}.    

DetectGPT \citep{mitchell2023detectgpt} evaluates a text $w_1\cdots w_T$ by first using a mask-filling model to generate perturbations $\tilde{w_1}\cdots \tilde{w_T}$ of the original text which still convey the same meaning but are differently expressed. They compute the mean log-likelihood $\overline{\mu}$ of these texts, along with the standard deviation $\sigma$, and assign the text a score $({P(w_1\cdots w_T)-\overline{\mu}})/{\sigma}$. Thus, rather than comparing the log-likelihood of $w_1\cdots w_T$ to a global threshold, they compare it to the log-likelihood of similar texts. DetectGPT marked a very significant upgrade to vanilla log-likelihood in the detection of machine-generated text. 

NPR \citep{su2023detectllm} translates the main idea of DetectGPT to log-rank. The method computes the ratio of the log-rank of $w_1\cdots w_T$ to the log rank of perturbations of the text, using the same perturbation model as DetectGPT.

Fast-DetectGPT (analytic version, see \cite{baofast} equation (5)) works the same way as DetectGPT except that, instead of using a mask-filling model, alternative samples are generated with the surrogate language model used for testing. In particular, $\tilde{\mu}=\frac{1}{T}\sum_{i=1}^T \sum_{v\in\mathcal V}p(v|w_{<i})\log(p(v|w_{<i}))$ here. As far as we understand, Fast-DetectGPT is currently the best-performing method for detecting machine-generated text. 

FourierGPT \citep{xu-etal-2024-detecting} and FACE \citep{yang2023face} are two promising techniques based on computing discrete Fourier transforms over sequences of log-likelihoods. These methods have the capability to extract subtle signals linked to periodicity, but also face a challenge in how to incorporate the computed spectra in a zero-shot classification test. FACE uses statistics such as Spectral Overlap or Spearman's correlation to compare spectra of collections of human and machine text at the distributional level, while FourierGPT forms a heuristic zero-shot test by summing the first $k$ coefficients, where $k$ is a parameter tuned over a validation set. Any such tuning can be considered training and may place these methods slightly outside the zero-shot regime.

Finally, Persistent Homology Dimension (PHD) \citep{tulchinskii2024intrinsic} evaluates a text by embedding it and computing the persistent homology dimension of the resulting embedded set. It seems to be very effective over long texts, yet the reasons for its effectiveness remain poorly understood. This technique is a notable exception in that it is not based on log-likelihood, log-rank or entropy.

\section{TEMPTEST: ZERO-SHOT MACHINE-GENERATED TEXT DETECTION}\label{sec:TempTest}

The denominators of equations \eqref{tempsampling} and \eqref{topksampling}, which are not independent of $w_1\cdots w_T$, represent unusual ways of normalizing probability. They normalize conditional next-token probabilities rather than normalizing the joint probability of a string. This is computationally very efficient but theoretically suboptimal, as discussed later. We term this effect {\it local normalization distortion} and study it in future work. One should expect this method of normalization to leave statistical traces that we can detect, and we empirically demonstrate this in Section \ref{sec:experiments}. In particular, these observations allow us to use the denominator of equation \eqref{tempsampling}, which we term TempNorm, as an additional metric to help assess whether a text has been generated using temperature sampling.

\begin{definition}\label{TempNorm}
The temperature-$\tau$ per-token $\log$-TempNorm of a sequence $\underline w=w_1\cdots w_T$ is given by
\[\frac{1}{T}\log\epsilon_{\tau}(\underline w)=\frac{1}{T}\sum_{i=1}^T\log\left(\sum_{v\in\mathcal V}(p(v|w_{<i})^{\frac{1}{\tau}})\right)
\]
\end{definition}

\begin{figure}[h]
    \vspace{.3in}
    \centering
    \includegraphics[width=0.45\textwidth]{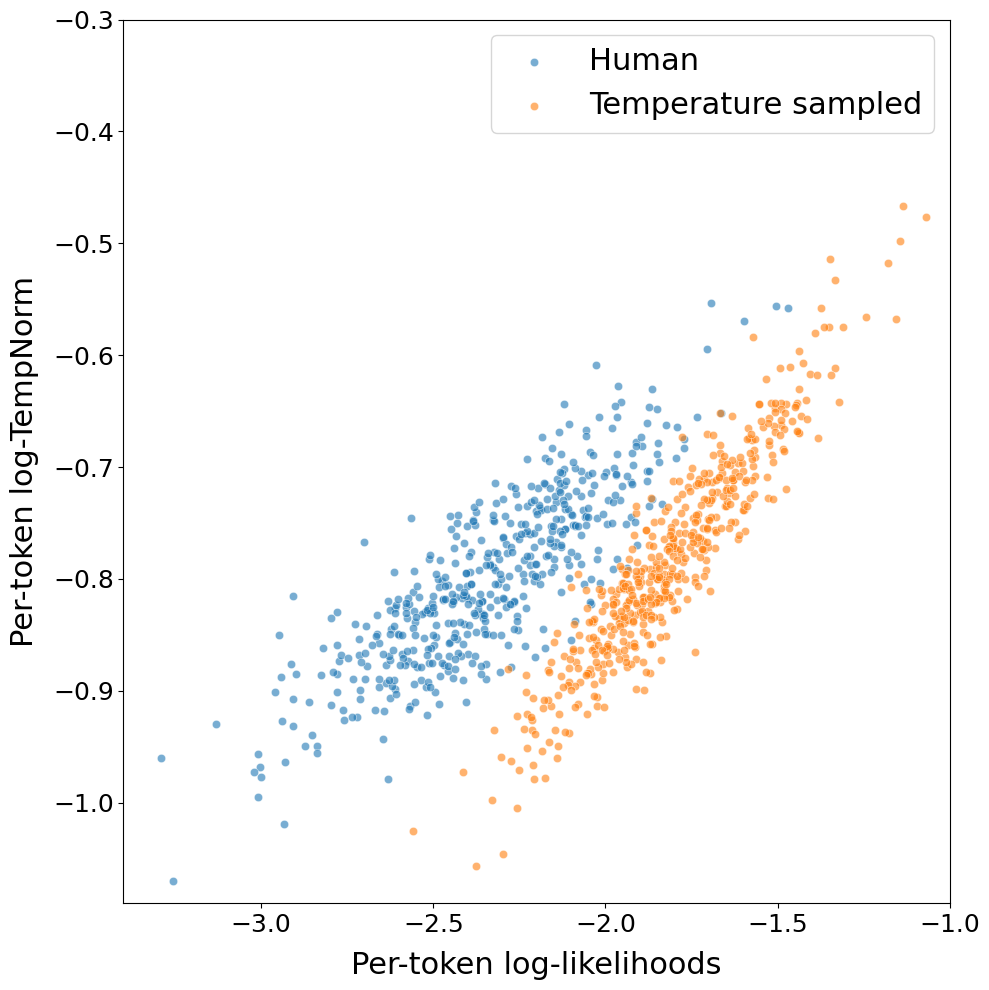}
    \vspace{.3in}
    \caption{Relationship Between Per-token Log-TempNorm and Per-token Log-likelihood. Observe that per-token log-likelihood alone does a poor job of distinguishing between these sets of 300-token human-written and Llama 3.1 temperature sampled text ($\tau=0.8$), as indicated by overlap in the orthogonal projection onto the $x$-axis. Adding a second coordinate displaying per-token log-TempNorm allows for linear separation of the texts.}
    \label{fig:demonstration}
\end{figure}

In Figure \ref{fig:demonstration} we plot the pair $$\left(\frac{1}{T}\log P(\underline w),\frac{1}{T}\log\epsilon_{\tau}(\underline w)\right)$$ 
and, by considering the orthogonal projection onto the x-axis, we see per token log-likelihood fails to effectively distinguish between the human and machine-generated texts. In contrast, adding in per-token log-TempNorm as the second coordinate makes it easy to draw a linear decision boundary dividing human texts and those generated by machine at temperature $\tau$. In TempTest we set the dividing line to have slope $\frac{1}{\tau}-1$ to construct a one-dimensional test. This value is theoretically justified in Section \ref{sec:Theory}.

\begin{definition}\label{TempTest} 
Given a sequence $\underline w=w_1\cdots w_T$, we define the TempTest score of $\underline w$ by
\[
\textnormal{TempTest}(\underline w):=\frac{1}{T}\left(\log\epsilon_{\tau}(\underline w)- \left(\frac{1}{\tau}-1\right)\log P(\underline w)\right).
\]
\end{definition}

In Section \ref{sec:Theory} we justify theoretically that, for sequences $\underline w_{\mbox{pure}}$ and $\underline w_{\tau}$ generated by pure-sampling and temperature $\tau$ sampling from $P$ respectively, one should expect that $\textnormal{TempTest}(\underline w_{\mbox{pure}}) > 0$ and $\textnormal{TempTest}(\underline w_{\tau}) < 0$. This is a strength compared to methods such as Fast-DetectGPT, where the statistic is not centered on a threshold of $0$.

We further show experimentally that the TempTest scores of human-written text are closer to those of pure-sampled text than temperature-sampled text and that TempTest is effective at distinguishing between human-written and temperature-sampled text. We stress that $0$ may not be the optimal threshold for distinguishing between human-written text and temperature-sampled text, particularly in the black box setting where a different language model is used for scoring, but empirically, we have found it to be a good estimate; for example, see Section \ref{sec:non-native}.

%%%%%%%%%%%%%%%%%%%%%%%%%%%%%%%%%%%%%%%%%%%%%%%%%%%
\subsection{Theoretical Justification}\label{sec:Theory}
%%%%%%%%%%%%%%%%%%%%%%%%%%%%%%%%%%%%%%%%%%%%%%%%%%%

We give two explanations as to why TempTest is effective at distinguishing between pure-sampled machine-generated text and temperature-sampled machine-generated text. First, a rigorous derivation from a Bayesian perspective, and second, an intuitive justification based on ergodic theory. These also explain our choice of parameter $\frac{1}{\tau}-1$ in Definition \ref{TempTest}. 

We do not have a theoretical explanation as to why TempTest scores for human-written text are closer to those of pure-sampled text than those of temperature-sampled text. Instead, we rely on the intuition that the manner in which humans generate text does not resemble temperature or top-k sampling, and so human-written text should not suffer from local normalization distortion. This hypothesis is supported by our experimental results, which show very clearly that TempTest is effective. 

%%%%%%%%%%%%%%%%%%%%%%%%%%%%%%%%%%%%%%%%%%%%%%%%%%%
\paragraph{A Bayesian Perspective}
%%%%%%%%%%%%%%%%%%%%%%%%%%%%%%%%%%%%%%%%%%%%%%%%%%%

Suppose we have a language model $P$ and flip a fair coin. If the outcome is heads, then we generate a text according to $P$ (pure-sampling); if it is tails, we generate a text $\underline w$ according to $Q_{\tau}$ (temperature sampling). Based only on knowledge of $\underline w$, the detection problem is equivalent to deciding whether we think the coin came down heads or tails.

Let $A$ be the event that the coin came down tails and the text was generated by $Q_{\tau}$. Let $B$ be the event that the generated text is $\underline w$. Bayes' rule tells us that
\[
\mathbb P(A|B)=\dfrac{\mathbb P(A)}{\mathbb P(B)} \mathbb P(B|A).
\]
We want to compute $\mathbb P(A|B)$, the probability that the generation method used temperature sampling, given the text. We know $\mathbb P(A)=\frac{1}{2}$.

$\mathbb P(B)$, the probability that our generation method produced text $\underline w$, satisfies 
\begin{align*}
    \mathbb P(B)&=\frac{1}{2}P(\underline w)+\frac{1}{2} Q(\underline w)\\
    &= \frac{1}{2}P(\underline w)+\frac{1}{2}\dfrac{P(\underline w)^{\frac{1}{\tau}}}{\epsilon_{\tau}(\underline w)}.
\end{align*}

Finally, $\mathbb P(B|A)=Q(\underline w)={P(\underline w)^{\frac{1}{\tau}}}/ {\epsilon_{\tau}(\underline w)}$.

So $\mathbb P(A|B)$, the probability that $\underline w$ was generated by temperature sampling, is given by
\begin{align*}
\mathbb P(A|B)&= \dfrac{1/2}{\frac{1}{2}\left(P(\underline w)+\frac{P(\underline w)^{\frac{1}{\tau}}}{\epsilon_{\tau}(\underline w)}\right)}\dfrac{P(\underline w)^{\frac{1}{\tau}}}{\epsilon_{\tau}(\underline w)}\\
&= \dfrac{1}{P(\underline w)^{1-\frac{1}{\tau}}\epsilon_{\tau}(\underline w)+1}.
\end{align*}
In particular, for a threshold $C\in(0,1)$, we conclude that $\mathbb P(A|B)>C$ if and only if
\[
\epsilon_{\tau}(\underline w)<\left(\frac{1}{C}-1\right)\frac{1}{P(\underline w)^{1-\frac{1}{\tau}}}.
\]
Using threshold $C={1}/{2}$, taking logs, and dividing by $T$, gives that $\mathbb P(A|B)>\frac{1}{2}$ if and only if
\[
\frac{1}{T}\log \epsilon_{\tau}(\underline w) <\left(\frac{1}{\tau}-1\right)\frac{1}{T}\log P(\underline w).
\]
This yields our formulation of TempTest.

%%%%%%%%%%%%%%%%%%%%%%%%%%%%%%%%%%%%%%%%%%%%%%%%%%%
\paragraph{Intuition From Ergodic Theory}
%%%%%%%%%%%%%%%%%%%%%%%%%%%%%%%%%%%%%%%%%%%%%%%%%%%

Given a value $\lambda$, what do texts generated by pure-sampling of per-token log-likelihood $\lambda$ look like? The mathematics is slightly cleaner if we consider the limit as passage length $T$ tends to infinity, and so consider the set
\[
\mathcal V_{\lambda}:=\{\underline w\in\mathcal V^{\mathbb N}: \lim_{T\to\infty}\frac{1}{T}\log p(w_1\cdots w_T)=\lambda\}.
\]
Since passages of equal $\log$-likelihood are equally likely under pure-sampling, we are really asking what the measure of maximal Kolmogorov-Sinai entropy on $\mathcal V_{\lambda}$ is. A fact well known to ergodic theorists and fractal geometers is that there exists a temperature $\tau$ such that this measure of maximal entropy is given by the Gibbs measure on $\mathcal V^{\mathbb N}$ at temperature $\tau$, see, for example, the section on the multifractal analysis of ergodic averages in \cite{falconer2014fractal}. This measure picks each string $w_1\cdots w_T$ with probability proportional to $p(w_1\cdots w_T)^{\frac{1}{\tau}}$. That is, these strings are distributed as those of temperature sampling if we were to remove the local normalization distortion. 

If we are then to ask whether a string of log-likelihood $\lambda$ looks more like pure-sampled strings of log-likelihood $\lambda$ or temperature $\tau$ sampled strings of log-likelihood $\lambda$, we are effectively asking about the local normalization distortion (TempNorm) of the string. This motivates the study of TempNorm and the development of TempTest.

\section{EXPERIMENTS}\label{sec:experiments}
%   - Summary of questions/hypothesis tested.
%   - Experimental setup for reproducibility
%   - Results sections
%%%%%%%%%%%%%%%%%%%%%%%%%%%%%%%%%%%%%%%%%%%%%%%%%%%%%%%%%%%%%%%%%%%%%%%%%%%%%%%%%%%%%%%%%%%%%%%%%%%%%%%%%%%%%%%%%%%%%%%%%%%%%%%%%%%%%%%%%%%%%%%%%%%%%%%%%

% Summary of questions/hypothesis tested.
According to TempTest, does human-written text look more like pure-sampled text or temperature-sampled text of the same log-likelihood?

To answer this question, we consider three core settings: white box, gray box, and black box. In the white box setting we have full access to the language model and temperature used to generate text and use this language model in the evaluation. In the gray box setting we have full access to the language model but do not know the temperature used in generation. In the black box setting we use a different language model for the evaluation of the text. For real-world applications it is important to assess fairness and robustness of new methods. Therefore, we test the degree of bias towards non-native speakers in TempTest, following a setup of \cite{tulchinskii2024intrinsic}, and robustness to paraphrasing attacks using Dipper \citep{krishna2024paraphrasing}.

%%%%%%%%%%%%%%%%%%%%%%%%%%%%%%%%%%%%%%%%%%%%%%%%%%%
\subsection{Setup} % Experimental setup for reproducibility
%%%%%%%%%%%%%%%%%%%%%%%%%%%%%%%%%%%%%%%%%%%%%%%%%%%

We follow \cite{mitchell2023detectgpt} by evaluating across three text datasets: news articles from XSum \citep{narayan2018don}, Wikipedia articles from SQuAD \citep{rajpurkar2016squad}, and stories from the Reddit WritingPrompts (Writing) dataset \citep{fan2018hierarchical}. In the black box setting, we also use generations by GPT3.5-Turbo and GPT-4 \citep{achiam2023gpt} supplied by Fast-DetectGPT \citep{baofast} and include the PubMedQA dataset \citep{jin2019pubmedqa}. Given a large language model, for each sample of human text we generated a synthetic version using the first $30$ tokens as context, except for PubMedQA where we take the whole question as context. To ensure a fair comparison, for both human and machine texts we then discarded the shared context to only consider the real and synthetic completions. Hereafter, sample size refers to the size of the completion, not including the original context.

For white box detection, we evaluate using language models Llama 3.1-8B \citep{dubey2024llama}, GPT2-XL \citep{radford2019language}, GPT-Neo-2.7B \citep{gpt-neo}, GPT-J-6B \citep{wang2021gpt} and OPT-2.7B \citep{zhang2022opt}. The choice of detection models largely follows \cite{baofast}.

All methods typically improve as the number of tokens in the sample to be tested increases. Indeed, state-of-the-art methods are nearing perfect AUROC scores at $200$ to $300$ tokens. Therefore, we primarily consider the most challenging scenario of small text snippets, typically using inputs of length $50$ tokens. This case highlights the strength of the signal and statistical power of the respective methods. For completeness, we include a comparison over longer texts too in Section \ref{sec:sample_size}.

As is standard, for example see \cite{mitchell2023detectgpt,baofast}, we use the AUROC measure to evaluate the performance of detectors. However, we note one practical strength of our method, not captured by this choice of metric, is a natural and theoretically justified decision threshold. Alongside recent methods detailed in Section \ref{sec:related_work}, we also use entropy, rank, and log-rank as baselines.

For all baseline methods, we have used the author's own implementations, drawing on the repository supplied with \cite{baofast}. Supporting code is available on GitHub\footnote{\url{https://github.com/TMKempton/TempTest}}. Experiments were run on a cluster of $8$ A100 GPUs.

\subsection{White Box}
Table \ref{tab:whitebox} details comparisons of TempTest against 6 baseline methods across 5 large language models and 3 datasets. TempTest was typically comparable or superior to the next best baseline with an average rank of $1.2$, compared to the Fast-DetectGPT which had an average rank of $1.6$.

\begin{table*}
\caption{White Box Experiment Results. In this setting the potential generation model and temperature are known at test time. Sample texts of length $50$ tokens were used, and $\tau=0.8$ for generation and scoring.}
\vspace{1em}
\label{tab:whitebox}
\resizebox{\linewidth}{!}{%
\begin{tabular}{@{}lllllllllllllllll@{}}
\toprule
Model          & \multicolumn{3}{c}{Llama 3.1}                                                      & \multicolumn{3}{c}{GPT2-XL}                                                        & \multicolumn{3}{c}{GPT-Neo 2.7b}                                                   & \multicolumn{3}{c}{GPT-J 6b}                                                       & \multicolumn{3}{c}{OPT-2.7b}                                                                 & \multicolumn{1}{c}{\multirow{2}{*}{Mean}} \\
Dataset        & \multicolumn{1}{c}{Writing} & \multicolumn{1}{c}{SQuAD} & \multicolumn{1}{c}{XSum} & \multicolumn{1}{c}{Writing} & \multicolumn{1}{c}{SQuAD} & \multicolumn{1}{c}{XSum} & \multicolumn{1}{c}{Writing} & \multicolumn{1}{c}{SQuAD} & \multicolumn{1}{c}{XSum} & \multicolumn{1}{c}{Writing} & \multicolumn{1}{c}{SQuAD} & \multicolumn{1}{c}{XSum} & \multicolumn{1}{c}{Writing} & \multicolumn{1}{c}{SQuAD} & \multicolumn{1}{c}{XSum}           & \multicolumn{1}{c}{}                      \\ \midrule
Entropy        & .516                        & .606                      & .573                     & .520                        & .616                      & .612                     & .508                        & .634                      & .624                     & .506                        & .645                      & .620                     & .516                        & .600                      & \multicolumn{1}{l|}{.569}          & .578                                      \\
Log-rank       & .815                        & .697                      & .683                     & .887                        & .862                      & .846                     & .883                        & .822                      & .838                     & .882                        & .780                      & .821                     & .887                        & .814                      & \multicolumn{1}{l|}{.842}          & .824                                      \\
Likelihood     & .806                        & .655                      & .662                     & .888                        & .839                      & .851                     & .885                        & .794                      & .844                     & .883                        & .747                      & .825                     & .879                        & .795                      & \multicolumn{1}{l|}{.842}          & .813                                      \\
DetectGPT      & .585                        & .498                      & .555                     & .798                        & .772                      & .781                     & .757                        & .717                      & .766                     & .708                        & .667                      & .702                     & .788                        & .676                      & \multicolumn{1}{l|}{.707}          & .698                                      \\
DetectLLM      & .607                        & .538                      & .593                     & .812                        & .538                      & .811                     & .774                        & .739                      & .787                     & .735                        & .721                      & .733                     & .816                        & .703                      & \multicolumn{1}{l|}{.731}          & .709                                      \\
Fast-DetectGPT & .914                        & .869                      & .790                     & .969                        & \textbf{.965}             & .966                     & \textbf{.963}               & \textbf{.954}             & \textbf{.963}            & .959                        & \textbf{.940}             & \textbf{.949}            & .959                        & .941                      & \multicolumn{1}{l|}{.945}          & .936                                      \\
\textbf{TempTest}      & \textbf{.935}               & \textbf{.893}             & \textbf{.848}            & \textbf{.972}               & .964                      & \textbf{.967}            & \textbf{.963}               & .951                      & \textbf{.963}            & \textbf{.960}               & .939                      & \textbf{.949}            & \textbf{.964}               & \textbf{.946}             & \multicolumn{1}{l|}{\textbf{.964}} & \textbf{.945}                             \\ \bottomrule
\end{tabular}
}
\end{table*}

\subsection{Gray Box}
\begin{figure}[t]
    \vspace{.3in}
    \centering
    \includegraphics[width=0.45\textwidth]{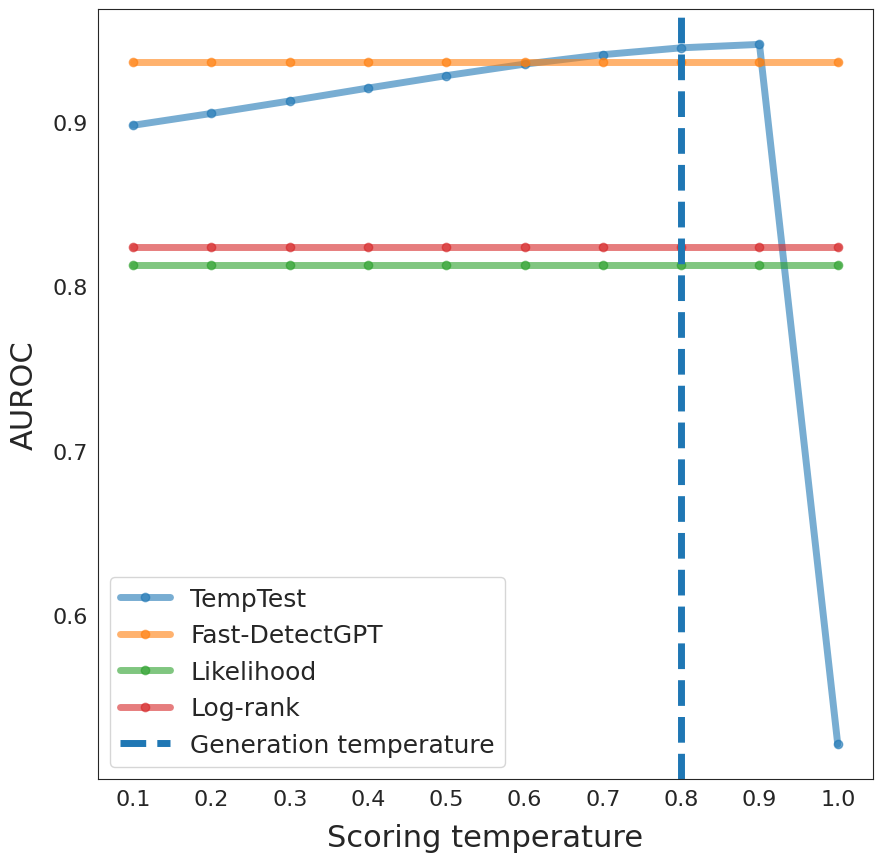}
    \vspace{.3in}
    \caption{Performance Across Scoring Temperatures. TempTest is robust to the particular choice of temperature used for scoring, even if it is significantly different from the ground-truth temperature used for generation. As expected, for $\tau=1$ the AUROC drops to $0.5$, as this case equivalent to pure-sampling. Meta Llama 3.1-8B is used for generation and scoring, and the ground-truth generation temperature is $0.8$.}\label{fig:greybox}
\end{figure}

Figure \ref{fig:greybox} demonstrates that TempTest is very robust to choosing the wrong scoring temperature. That is, if the generation temperature is not known, as in a realistic scenario, choosing arbitrarily should suffice. For example, if auditing a selection of creative writing essays, an educated guess might suppose that counterfeits are using a relatively high temperature such as $0.7$ to $0.8$ for greater generated diversity.

\subsection{Black Box}\label{subsec:blackbox}
\begin{figure}[t]
    \vspace{.3in}
    \centering
    \includegraphics[width=0.45\textwidth]{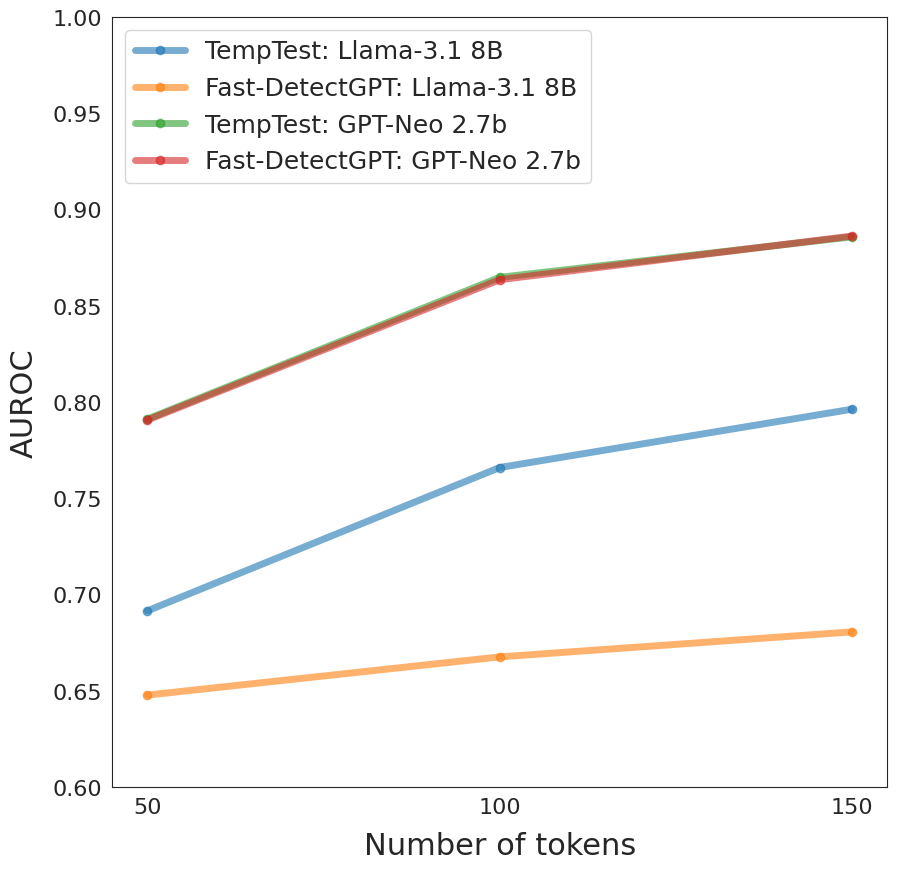}
    \vspace{.3in}
    \caption{Performance On GPT-3.5 Turbo And GPT-4 Generated Data. TempTest obtains comparable or superior performance to Fast-DetectGPT across a range of passage lengths in an illustrative black-box setting when using GPT-3.5 Turbo or GPT-4 as the generation model. Two scoring models were considered here: Meta Llama 3.1-8B and GPT-Neo 2.7b. A temperature of $0.8$ was used for generation and scoring. Results are averaged over Writing, XSum, and PubMedQA completions.}\label{fig:gpt}
\end{figure}

Table \ref{tab:blackbox} details comparisons of TempTest against 6 baseline methods across $4$ large language models and $3$ datasets. TempTest was typically comparable or superior to the next best baseline with an average rank of $1.3$, compared to Fast-DetectGPT, which had an average rank of $1.6$. In addition, in Figure \ref{fig:gpt} we show TempTest is stronger than or equal to Fast-DetectGPT over 3 sample sizes on synthetic data from GPT-3.5-Turbo and GPT-4. We also benchmarked PHD \citep{tulchinskii2024intrinsic}, and found it to be comparable with middle-ranked baselines, see Appendix \ref{app:phd}.

\begin{table*}
\caption{Black Box Experiment Results. In this setting the temperature is known, but the generation model, Meta Llama 3.1-8B, is not known at test time. Hence, a variety of alternative scoring models were evaluated. Sample texts of length $50$ tokens were used. If the temperature is also unknown and must be estimated, only a small degradation is expected; see Figure \ref{fig:greybox}.}
\vspace{1em}
\label{tab:blackbox}
\resizebox{\linewidth}{!}{%
\begin{tabular}{@{}llllllllllllll@{}}
\toprule
Scoring Model             & \multicolumn{3}{c}{GPT2-XL}                                                        & \multicolumn{3}{c}{GPT-Neo 2.7b}                                                   & \multicolumn{3}{c}{GPT-J 6b}                                                       & \multicolumn{3}{c}{OPT-2.7b}                                                                 & \multicolumn{1}{c}{\multirow{2}{*}{Mean}} \\
Dataset           & \multicolumn{1}{c}{Writing} & \multicolumn{1}{c}{SQuAD} & \multicolumn{1}{c}{XSum} & \multicolumn{1}{c}{Writing} & \multicolumn{1}{c}{SQuAD} & \multicolumn{1}{c}{XSum} & \multicolumn{1}{c}{Writing} & \multicolumn{1}{c}{SQuAD} & \multicolumn{1}{c}{XSum} & \multicolumn{1}{c}{Writing} & \multicolumn{1}{c}{SQuAD} & \multicolumn{1}{c}{XSum}           & \multicolumn{1}{c}{}                      \\ \midrule
Entropy           & .492                        & .538                      & .567                     & .478                        & .527                      & .524                     & .492                        & .560                      & .549                     & .478                        & .533                      & \multicolumn{1}{l|}{.526}          & .522                                      \\
Log-rank          & .802                        & .722                      & .622                     & .807                        & .733                      & .702                     & .807                        & .715                      & .692                     & .807                        & .727                      & \multicolumn{1}{l|}{.694}          & .736                                      \\
Likelihood        & .770                        & .686                      & .562                     & .794                        & .707                      & .684                     & .793                        & .681                      & .670                     & .800                        & .711                      & \multicolumn{1}{l|}{.690}          & .712                                      \\
DetectGPT         & .575                        & .566                      & .513                     & .587                        & .553                      & .567                     & .572                        & .508                      & .548                     & .566                        & .561                      & \multicolumn{1}{l|}{.583}          & .558                                      \\
DetectLLM         & .628                        & .584                      & .563                     & .619                        & .561                      & .584                     & .600                        & .544                      & .586                     & .579                        & .556                      & \multicolumn{1}{l|}{.592}          & .583                                      \\
Fast-DetectGPT    & \textbf{.847}               & \textbf{.797}             & .643                     & \textbf{.880}               & .825                      & .754                     & \textbf{.882}               & \textbf{.827}             & .755                     & .887                        & .837                      & \multicolumn{1}{l|}{.775}          & .809                                      \\
\textbf{TempTest} & \textbf{.847}               & .794                      & \textbf{.647}            & .877                        & \textbf{.826}             & \textbf{.760}            & .881                        & .822                      & \textbf{.758}            & \textbf{.898}               & \textbf{.843}             & \multicolumn{1}{l|}{\textbf{.831}} & \textbf{.815}                             \\ \bottomrule
\end{tabular}
}
\end{table*}

\subsection{Effect Of Sample Size}\label{sec:sample_size}

\begin{figure}
    \vspace{.3in}
    \centering
    \includegraphics[width=0.45\textwidth]{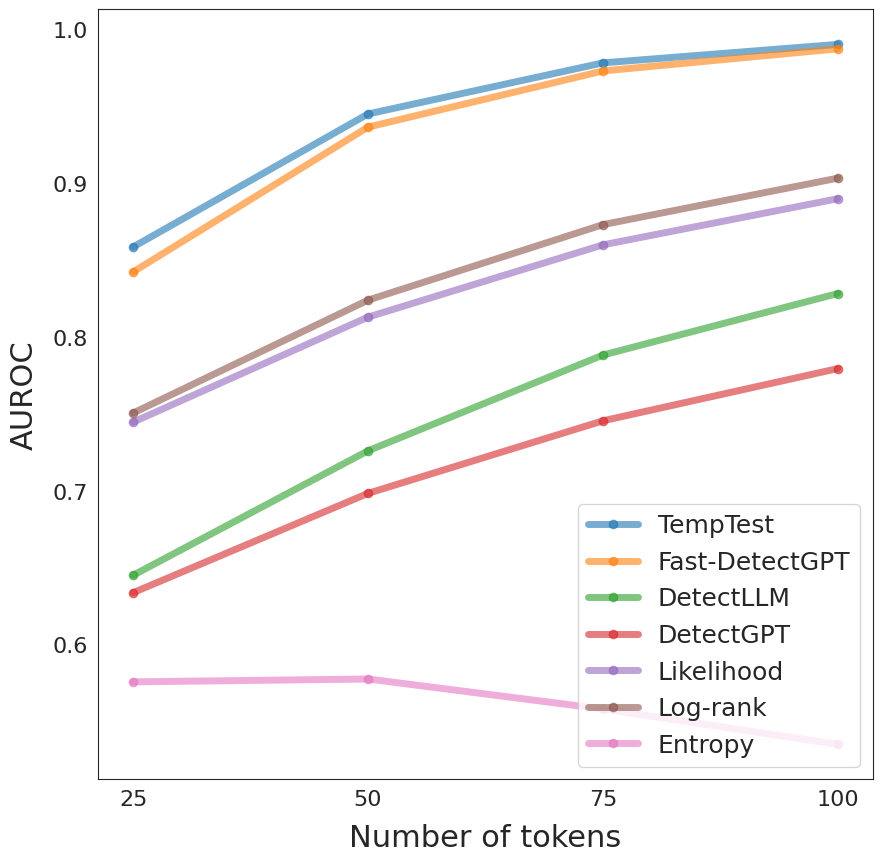}
    \vspace{.3in}
    \caption{Performance For Varying Context Lengths. AUROCs averaged across all 3 datasets and 5 models used in Table \ref{tab:whitebox} at input text sizes ranging from $25$ to $100$ tokens.}\label{fig:sample-size}
\end{figure}

Our white and black box experiments focus on a sample size of $50$, a challenging task that highlights statistical power and helps to separate methods. As sample size increases, we typically see all methods perform better, and ultimately converge. Figure \ref{fig:sample-size} shows AUROCs averaged across all datasets and models used in Table \ref{tab:whitebox}. TempTest outperforms all baselines at low sample sizes and is marginally better or comparable at higher sample sizes.

\subsection{Non-native Speaker Bias}\label{sec:non-native}

Machine-generated text detection algorithms have been shown to be potentially biased against non-native speakers by classifying them as machines \citep{Liang2023GPTDA}. 
To evaluate TempTest in this setting, we followed the broad setup of \cite{Liang2023GPTDA}. This involved comparing detection algorithms on TOEFL essays written by non-native speakers with the same essays re-written by GPT-4. GPT-4 was prompted to re-write the essays to sound more like a non-native speaker, see Appendix \ref{app:exp-sum} for details. To obtain concrete predictions from scores, \cite{Liang2023GPTDA} use decision thresholds determined by the \emph{equal error rate} (EER) on a held-out dataset (Writing \citep{fan2018hierarchical}). For both tuning and scoring we use GPT-Neo 2.7b, the favored model of Fast-DetectGPT. Figure \ref{fig:non-native} (Appendix \ref{subsec:additionalresults}) shows TempTest has improved overall error rates to the baseline Fast-DetectGPT, and TempTest was less likely to classify the human text as machine-generated. This preliminary result suggests TempTest may be effective in reducing some sources of bias in machine-generated text detection. Although we roughly followed the experimental setup of \cite{tulchinskii2024intrinsic} for continuity with the existing literature, it should be noted that this is a small-scale experiment due to the limited available data. We encourage the development of larger datasets, new methods to mitigate bias, and more extensive analysis in a range of experimental settings.

\subsection{Robustness To Paraphrasing Attacks}

\begin{figure}
    \vspace{.3in}
    \centering
    \includegraphics[width=0.45\textwidth]{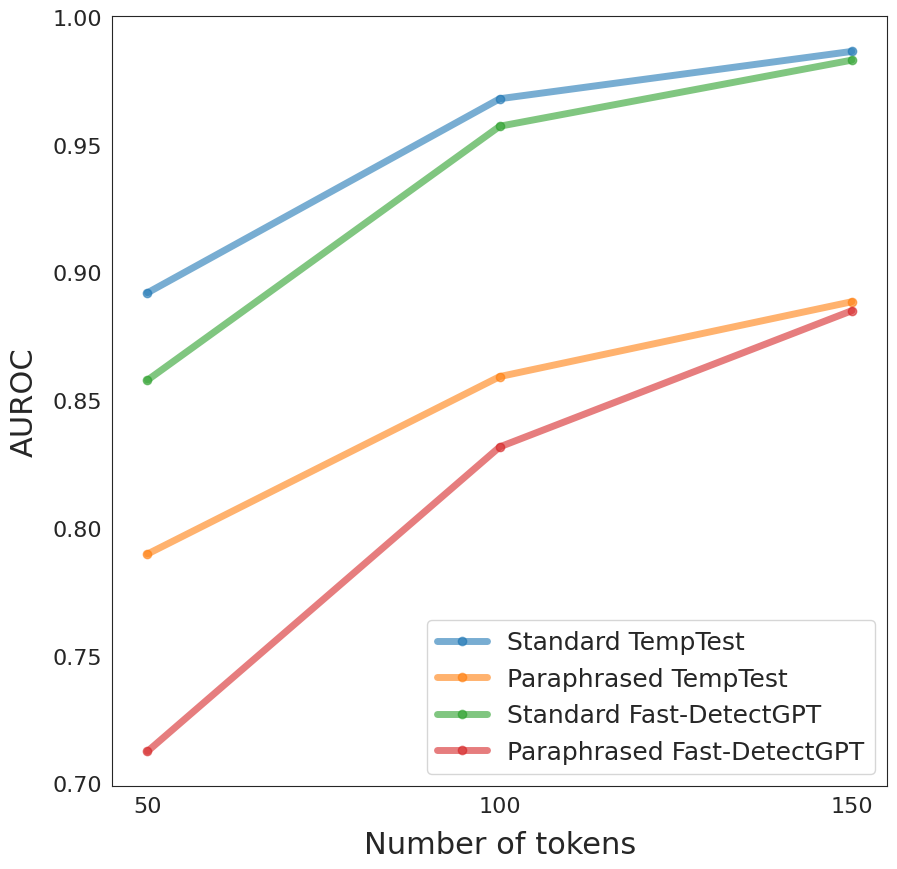}
    \vspace{.3in}
    \caption{Robustness Of TempTest Under Paraphrasing. TempTest is less susceptible to paraphrasing attacks than the previous art, Fast-DetectGPT, over a range of input text sizes. Reported AUROC values are the average over datasets Writing, XSum, and SQuAD.}\label{fig:dipper}
\end{figure}

Paraphrasing attacks, using models such as Dipper \citep{krishna2024paraphrasing}, have been shown to significantly degrade the performance of machine-generated text detectors. Figure \ref{fig:dipper} shows that TempTest is impacted by such attacks, but is far less susceptible than the best baseline, Fast-DetectGPT. A wider suite of adversarial attacks is investigated in \citep{Liu2024OnET}, and we leave it to future work to verify the effectiveness of TempTest in this context.

\section{DISCUSSION}

Section \ref{sec:experiments} empirically validates the effectiveness of TempTest as a zero-shot method for detecting machine-generated text. However, it is reasonable to ask how TempTest and other methods of detecting machine-generated text will fare in the future as language models become more sophisticated. One may expect that methods based on log-likelihood or log-rank will see their performance degrade as the probability distributions that language models output start to resemble human texts more closely, a hypothesis foreshadowed by our results on Llama 3.1. Similarly, we would expect that, when plotting log-likelihood against TempNorm for pure-sampled machine-generated texts, they will start to resemble the plots for human texts more closely. However, the statistical differences between pure-sampled text and temperature-sampled text are not a function of the quality of the language model, and so one should not expect TempTest to become less good at differentiating between human written and temperature-sampled text.  Indeed, the main threat to the longevity of our method is that temperature sampling and top-k sampling may go out of fashion.

\section{CONCLUSIONS}
Across a wide range of scenarios, we have demonstrated that TempTest is an effective new approach for detecting machine-generated text that is comparable or superior to the current state-of-the-art, with this effect most pronounced for short passage lengths and on modern language models. This test is based on a novel statistical quantity, TempNorm, which quantifies the local normalization distortion of conditional probability measures. Importantly, this provides an extra tool for researchers and practitioners to use when building future detection algorithms.

However, there are drawbacks to our method that present opportunities for future work. First, we consider only English and these empirical investigations should be expanded to include other languages. Second, the distortion we detect is only present in temperature-sampled text and thus is not applicable to text generated through different generation strategies such as top-p or top-k. That said, an analogous method may be derived for top-k, see Appendix \ref{app:top-k}. Moreover, we hope that future work will look at ways of ensembling or unifying different zero-shot methods into a single approach, creating methods that are broadly applicable and exploiting the advantages of each technique. Ultimately, the more diverse our statistical toolbox is for analyzing text, the better we will be able to detect the traces machines leave behind.

\subsubsection*{Acknowledgements}
Computational resources were provided by Featurespace. We also thank our reviewers for their helpful comments.

\bibliography{content/references}

\onecolumn
\aistatstitle{SUPPLEMENTARY MATERIALS}

\section{EXPERIMENTAL DETAILS}\label{app:exp-sum}
We gather here experimental details described elsewhere in the paper. 

Usually, when generating machine text, 30 tokens of human text are given as a prompt to the language model, which generates a completion. The exception is the PubMedQA dataset \citep{jin2019pubmedqa}, which contains questions and answers. In that case, the whole question was given as a prompt to the language model. Machine-generated texts were of length 50 tokens except where stated otherwise, for example Figure \ref{fig:sample-size}.

\paragraph{White Box} In these experiments, machine-generated texts were generated using the XSum, SQuAD and Writing datasets. The machine-generated texts were generated by Llama 3.1-8B, GPT2-XL, GPT-Neo-2.7B, GPT-J-6B and OPT-2.7B, each at temperature $0.8$. In each case, the same language model was used for evaluation, and TempTest was run with the correct temperature of $0.8$.

\paragraph{Black Box} We ran two separate experiments. In the first, texts were generated using GPT2-XL, GPT-Neo-2.7B, GPT-J-6B and OPT-2.7B at temperature $0.8$, with prompts from XSum, SQuAD and Writing. The texts were evaluated using Llama 3.1-8B. For the second set of experiments (Tables \ref{tab:gpt-neo} and \ref{tab:gpt-llama}), texts were generated using GPT3.5 Turbo and GPT4 at temperature $0.8$, with prompts from Writing, PubMedQA and XSum. Llama 3.1-8B and GPT-Neo-2.7B were used as evaluators; see Subsection \ref{subsec:comment} for the rationale behind these scoring models. 

\paragraph{Gray box} We experimented with the case where the generating model is known but the generating temperature is not. In this case, we generated texts using Llama 3.1-8B at temperature $0.8$, and ran TempTest with scoring model Llama 3.1-8B at a variety of temperatures.

\subsection{A Comment On The Choice of Scoring Model}\label{subsec:comment}
The datasets (XSum, PubMedQA, Writing and SQuAD) upon which we evaluate detector performance have been used in many previous works, including state-of-the-art detector FastDetect-GPT. These works evaluated their detectors using a range of different language models as scorers. In order to avoid selecting the scoring language model based on performance on the test set, we took the decision to use Llama 3.1-8B as the primary black-box scoring model, since this is a modern language model which had not previously been used with existing art. The one exception is in our experiment on texts generated by GPT-3.5 Turbo and GPT4. Here, it appears that FastDetect-GPT performs much worse when Llama is used as the detection model, and so we also included results using the best-performing language model for their paper, GPT-Neo 2.7B. See Subsection \ref{subsec:additionalresults}.
%\vfill

\subsection{Two Cautionary Tales}\label{subsec:cautionary}
We mention two things to be aware of when evaluating the performance of detectors. The first is that, at time of writing, the HuggingFace \texttt{GenerationConfig} class has top-k set to 50 by default, and so one must explicitly set top-k to $0$ or \texttt{None} to deactivate this. At least three papers in the area have reported results for detecting pure-sampled text where the experiment was actually run on top-k ($k=50$) sampled text as a result of this choice of default. The second thing to be aware of is that several detectors produce results whose expected values are not independent of the length of the text. This is not in itself a problem, provided calibration of the correct classification threshold is done separately for each value of text length $T$. It does, however, mean that AUROC values computed over datasets where the human and machine-generated text have different lengths can be highly misleading. Some papers in the area study machine texts of 200 or 300 tokens, whereas the human written texts vary naturally in length but are typically much lower, which can heavily skew the results.  

\section{ADDITIONAL RESULTS AND METHODS}\label{subsec:additionalresults}
\subsection{Further experiments on GPT-3.5 and GPT-4}
In Tables \ref{tab:gpt-neo} and \ref{tab:gpt-llama} we present the full (disaggregated) results used in Figure \ref{fig:gpt}, where we compared the performance of TempTest to FastDetect-GPT for detecting text generated by GPT-3.5 Turbo and GPT-4 at temperature 0.8.

\begin{table}[t]
\caption{Detecting GPT-3.5 And GPT-4 Generated Text With Scoring Model GPT-Neo 2.7b. In this experiment the generation temperature was $0.8$ and the text length was 50 tokens. TempTest was evaluated with scoring temperatures, denoted in parentheses, between $0.6$ and $0.9$.}
\vspace{1em}
\label{tab:gpt-neo}
\centering
\begin{tabular}{@{}llllllll@{}}
\toprule
Generation Model                   & \multicolumn{3}{c}{GPT-3.5 Turbo}                                                   & \multicolumn{3}{c}{GPT-4}                                                                     & \multicolumn{1}{c}{\multirow{2}{*}{Mean}} \\
Dataset                 & \multicolumn{1}{c}{Writing} & \multicolumn{1}{c}{PubMedQA} & \multicolumn{1}{c}{XSum} & \multicolumn{1}{c}{Writing} & \multicolumn{1}{c}{PubMedQA} & \multicolumn{1}{c}{XSum}           & \multicolumn{1}{c}{}                      \\ \midrule
Fast-DetectGPT          & .861                        & .785                       & \textbf{.895}            & .727                        & .694                       & \multicolumn{1}{l|}{\textbf{.783}} & \textbf{.791}                             \\
TempTest ($\tau = 0.6$) & .861                        & \textbf{.815}              & .878                     & .704                        & \textbf{.713}              & \multicolumn{1}{l|}{.741}          & .785                                      \\
TempTest ($\tau = 0.7$) & .863                        & .809                       & .883                     & .715                        & .708                       & \multicolumn{1}{l|}{.754}          & .789                                      \\
TempTest ($\tau = 0.8$) & \textbf{.866}               & .800                       & .889                     & .725                        & .701                       & \multicolumn{1}{l|}{.767}          & \textbf{.791}                             \\
TempTest ($\tau = 0.9$) & \textbf{.866}               & .785                       & .893                     & \textbf{.733}               & .694                       & \multicolumn{1}{l|}{.777}          & \textbf{.791}                             \\ \bottomrule
\end{tabular}
\end{table}

\begin{table}[t]
\caption{Detecting GPT-3.5 And GPT-4 Generated Text With Scoring Model Meta Llama 3.1-8B. In this experiment the generation temperature was $0.8$ and the text length was 50 tokens. TempTest was evaluated with scoring temperatures, denoted in parentheses, between $0.6$ and $0.9$.}
\vspace{1em}
\label{tab:gpt-llama}
\centering
\begin{tabular}{@{}llllllll@{}}
\toprule
Generation Model                   & \multicolumn{3}{c}{GPT-3.5 Turbo}                                                   & \multicolumn{3}{c}{GPT-4}                                                                     & \multicolumn{1}{c}{\multirow{2}{*}{Mean}} \\
Dataset                 & \multicolumn{1}{c}{Writing} & \multicolumn{1}{c}{PubMedQA} & \multicolumn{1}{c}{XSum} & \multicolumn{1}{c}{Writing} & \multicolumn{1}{c}{PubMedQA} & \multicolumn{1}{c}{XSum}           & \multicolumn{1}{c}{}                      \\ \midrule
Fast-DetectGPT          & .705                        & .738                       & .627                     & \textbf{.698}               & .593                       & \multicolumn{1}{l|}{.526}          & .648                                      \\
TempTest ($\tau = 0.6$) & \textbf{.796}               & \textbf{.812}              & \textbf{.732}            & .697                        & \textbf{.667}              & \multicolumn{1}{l|}{.570}          & \textbf{.712}                             \\
TempTest ($\tau = 0.7$) & .777                        & .798                       & .725                     & .696                        & .650                       & \multicolumn{1}{l|}{.574}          & .703                                      \\
TempTest ($\tau = 0.8$) & .755                        & .779                       & .715                     & .693                        & .631                       & \multicolumn{1}{l|}{\textbf{.577}} & .692                                      \\
TempTest ($\tau = 0.9$) & .726                        & .757                       & .700                     & .689                        & .611                       & \multicolumn{1}{l|}{\textbf{.577}} & .677                                      \\ \bottomrule
\end{tabular}
\end{table}

%%%%%%%%%%%%%%%%%%%%%%%%%%%%%%%%%%%%%%%%%%%%%%%%%%%
\subsection{Detecting Top-k Generated Text}\label{app:top-k}
%%%%%%%%%%%%%%%%%%%%%%%%%%%%%%%%%%%%%%%%%%%%%%%%%%%

We describe a test analogous to TempTest for detecting text which has been generated by a language model using top-k sampling.

The simplest white box method for detecting whether a text, or a portion of a text, has been generated by a given language model using top-k sampling is to scan through the text, looking for passages where each token lies in the top-k set. In \cite{gehrmann2019gltr} this was used to produce a color overlay for passages of text, where tokens in the top-10 set were colored green, tokens in the top-100 set were colored yellow and tokens in the top-1000 set were colored red.

As always, one runs the risk of false positives. Scanning through human-written texts from the writing prompts dataset, we find $61$ of the $500$ essays contain at least one passage of at least $30$ consecutive tokens, which lie in the top-50 set. This begs the question of whether one can distinguish between human-written passages which lie in the top-50 set and passages generated by top-50 sampling.  

Our method follows the reasoning of TempTest. Recall that the Bayesian justification for TempTest was not able to compare human-written text with temperature-sampled text; instead there was a rigorously justified method for assessing whether a text was more likely generated by pure sampling or temperature sampling, and then experimental evidence that this method was also effective at distinguishing between human-written and temperature-sampled text. Similarly, given a passage of text that lies in the top-k set, we give a rigorous justification of a method to determine whether it was more likely generated by top-k sampling or pure sampling.

We recall that the denominators in equation \eqref{topksampling} are non-constant, and so top-k sampling does not sample strings $w_1\cdots w_T$ from the top-k set with probability proportional to $p(w_1\cdots w_T)$. This means that passages $w_i\cdots w_l$ for which, on average, the top-k set is rather small are (relatively) much more likely under top-k sampling than they would have been by sampling directly from $p$. 

Another way of viewing this is that it is much easier to accidentally generate a passage of text for which each token lies in the corresponding top-50 set when these top-50 sets have a large mass. We reproduce below the argument of Section \ref{sec:Theory} for a careful Bayesian justification of this claim. This leads us to the following:

\paragraph{Previous test (in the spirit of \cite{gehrmann2019gltr})} Scan the text to identify passages of at least thirty consecutive tokens lying in the top-50 set. Declare such passages suspicious

\paragraph{New test} Scan the text to identify passages of at least thirty consecutive tokens lying in the top-50 set. Given such a passage $w_j\cdots w_l$, compute the geometric mean of the size of top-50 sets arising from the passage, i.e. 
\[\exp\left(\frac{1}{l-j+1}\log\left(\prod_{i=j}^l \sum_{v\in\mathcal V_k(w_{<i})}p(v|w_{<i})\right)\right). 
\]
If this mean size is below some threshold $C$, declare the passage suspicious.

\begin{figure}
    \vspace{.3in}
    \centering
    \includegraphics[width=0.45\textwidth]{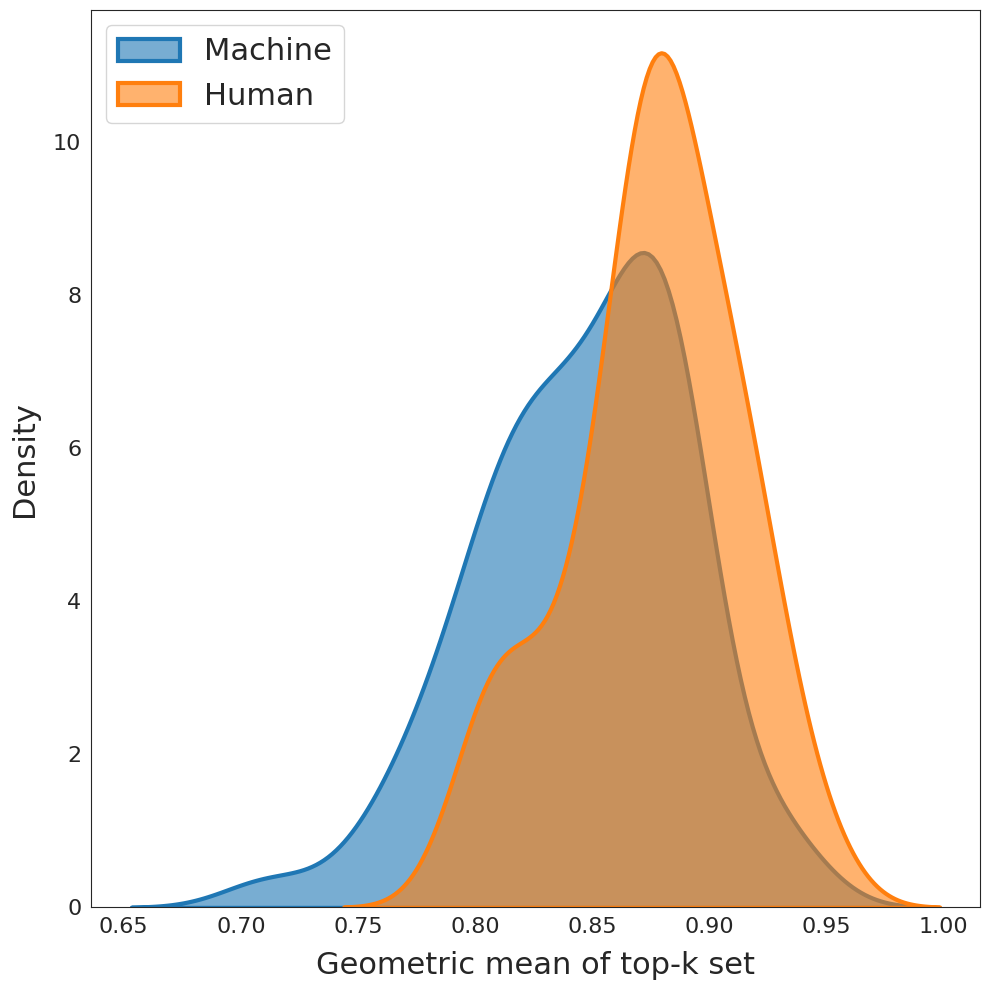}
    \vspace{.3in}
    \caption{Comparing Geometric Means Of Machine And Human Text. This density plot highlights a discrepancy between the geometric mean of the size of the top-k set for human-written passages that are coincidentally in the top-k set and machine-generated passages that used top-k sampling.}
    \label{fig:your_label}
\end{figure}

We apply these tests to human-written texts in the Writing dataset. Wherever a human-written text $\underline w$ has a passage $w_i\cdots w_j$ of at least thirty tokens in the top-50 set, we use context $w_1\cdots w_{i-1}$ to prompt Llama 3.1 with top-50 sampling to generate an alternative passage of the same length. We then compare in Figure \ref{fig:your_label} the geometric mean of the size of the top-50 set across these human and machine passages.  

Our new test achieves an AUROC of 0.695 in distinguishing human-written passages, which would have been declared suspicious by the old test from top-50 sampling machine-generated passages. Thus, even on passages of length thirty tokens, the extra information added by top-k-mass significantly increases our ability to reliably detect passages generated by top-k sampling.

\subsubsection{The Bayesian Perspective On Our Top-k Test}
Let us fix a language model $P$ and a natural number $k$. We wish to compare the probabilities that a text was generated by pure sampling or top-k sampling, conditioned on the fact that this text is known to lie in the top-k set. Equivalently, we are asking for the relative probabilities that the text was generated by top-k sampling or rejection sampling, where the rejection sampling comes from pure sampling repeatedly until one generates a sample which lies in the top-k set. This rejection sampling assigns mass $P(\underline w)/C$ to elements of the top-k set, where $C$ is some unknown constant.

Suppose that we flip a fair coin and according to the outcome we generate a text according to rejection-sampling $P/C$ (if the coin came down heads) or top-k sampling $Q_k$ (if the coin came down tails). Suppose we are given a text $\underline w=w_1\cdots w_T$ which lies in the top-k set. Based only on knowledge of $\underline w$, the detection problem is equivalent to deciding whether we think the coin came down heads or tails.

Let $A$ be the event that the coin came down tails and the text was generated by top-k sampling. Let $B$ be the event that the generated text is $\underline w$. Bayes' rule tells us that
\[
\mathbb P(A|B)=\dfrac{\mathbb P(A)}{\mathbb P(B)} \mathbb P(B|A).
\]
We want to compute $\mathbb P(A|B)$, the probability that the generation method used top-k sampling, given the text. We know $\mathbb P(A)=\frac{1}{2}$.

$\mathbb P(B)$, the probability that our generation method produced text $\underline w$, satisfies 
\begin{align*}
    \mathbb P(B)&=\frac{1}{2C}P(\underline w)+\frac{1}{2} Q_k(\underline w)\\
    &= \frac{1}{2C}P(\underline w)+\frac{1}{2}\dfrac{P(\underline w)}{\epsilon_{k}(\underline w)}
\end{align*}

where \[\epsilon_k(\underline w):=\prod_{i=1}^T\sum_{v\in \mathcal V_k(w_{<i})} p(v|w_{<i}) \]
is the product of the masses of the top-k sets at each time $i\in\{1,\cdots T\}$. Finally, $\mathbb P(B|A)=Q_k(\underline w)={P(\underline w)}/{\epsilon_{k}(\underline w)}$.

So $\mathbb P(A|B)$, the probability that $\underline w$ was generated by top-k sampling, is given by
\begin{align*}
\mathbb P(A|B)&= \dfrac{1/2}{\frac{1}{2}\left(\frac{P(\underline w)}{C}+\frac{P(\underline w)}{\epsilon_{k}(\underline w)}\right)}\dfrac{P(\underline w)}{\epsilon_{k}(\underline w)}\\
&= \dfrac{1}{1+\frac{\epsilon_{k}(\underline w)}{C}}.
\end{align*}
In particular, since $C$ is a fixed constant independent of $\underline w$, we see that the conditional probability that the text was generated by top-k sampling depends only on the product of the masses of the top-k sets along the sequence $\underline w$, as required. 
\subsection{PHD Results}\label{app:phd}
\begin{figure}
    \vspace{.3in}
    \centering
    \includegraphics[width=0.45\textwidth]{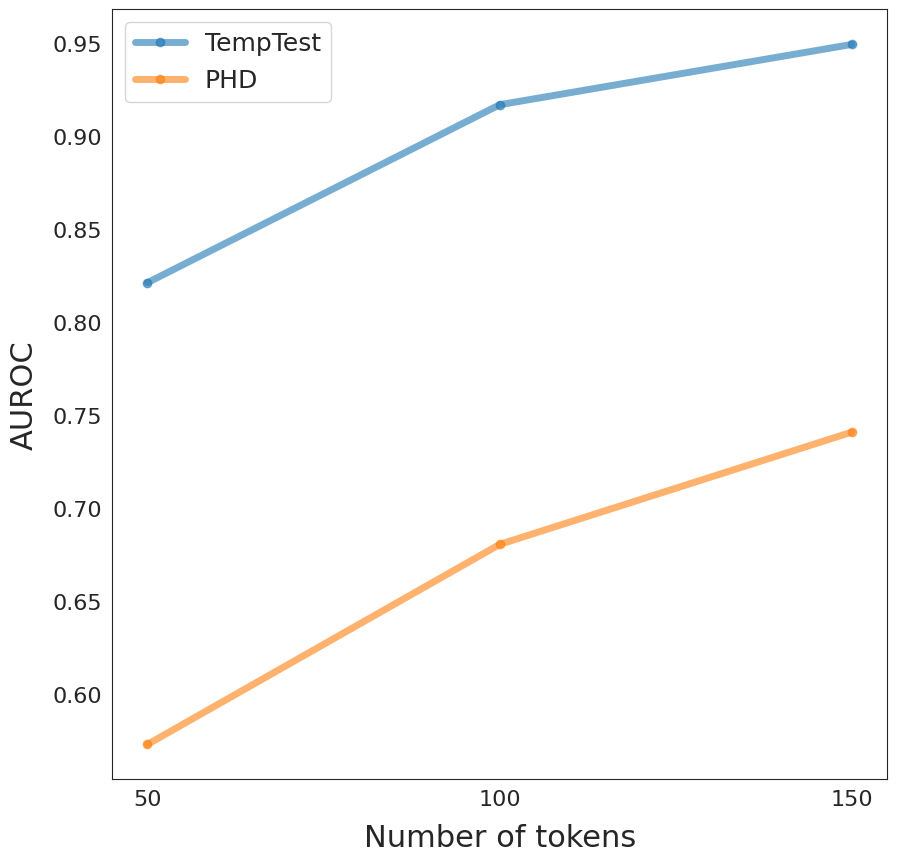}
    \vspace{.3in}
    \caption{PHD And TempTest In A Black Box Setting. Roberta-base is used for PHD as the embedding model as recommended in \cite{tulchinskii2024intrinsic}, while GPT-Neo 2.7B is used for scoring with TempTest. Temperature $0.8$ was used throughout. Results are aggregated over Writing, XSum, and SQuAD.}
    \label{fig:PHD}
\end{figure}

As mentioned in Subsection \ref{subsec:blackbox}, we benchmark TempTest against PHD, a method for detecting machine-generated text based on persistent homology dimension \cite{tulchinskii2024intrinsic}; see Figure \ref{fig:PHD}. Results are always in the black box setting, since PHD uses an embedding model for evaluation rather than a standard language model. Results are aggregated over Writing, XSum, and SQuAD.

\subsection{Experiment On TOEFL Data}
\begin{figure}
    \vspace{.3in}
    \centering
    \includegraphics[width=0.8\textwidth]{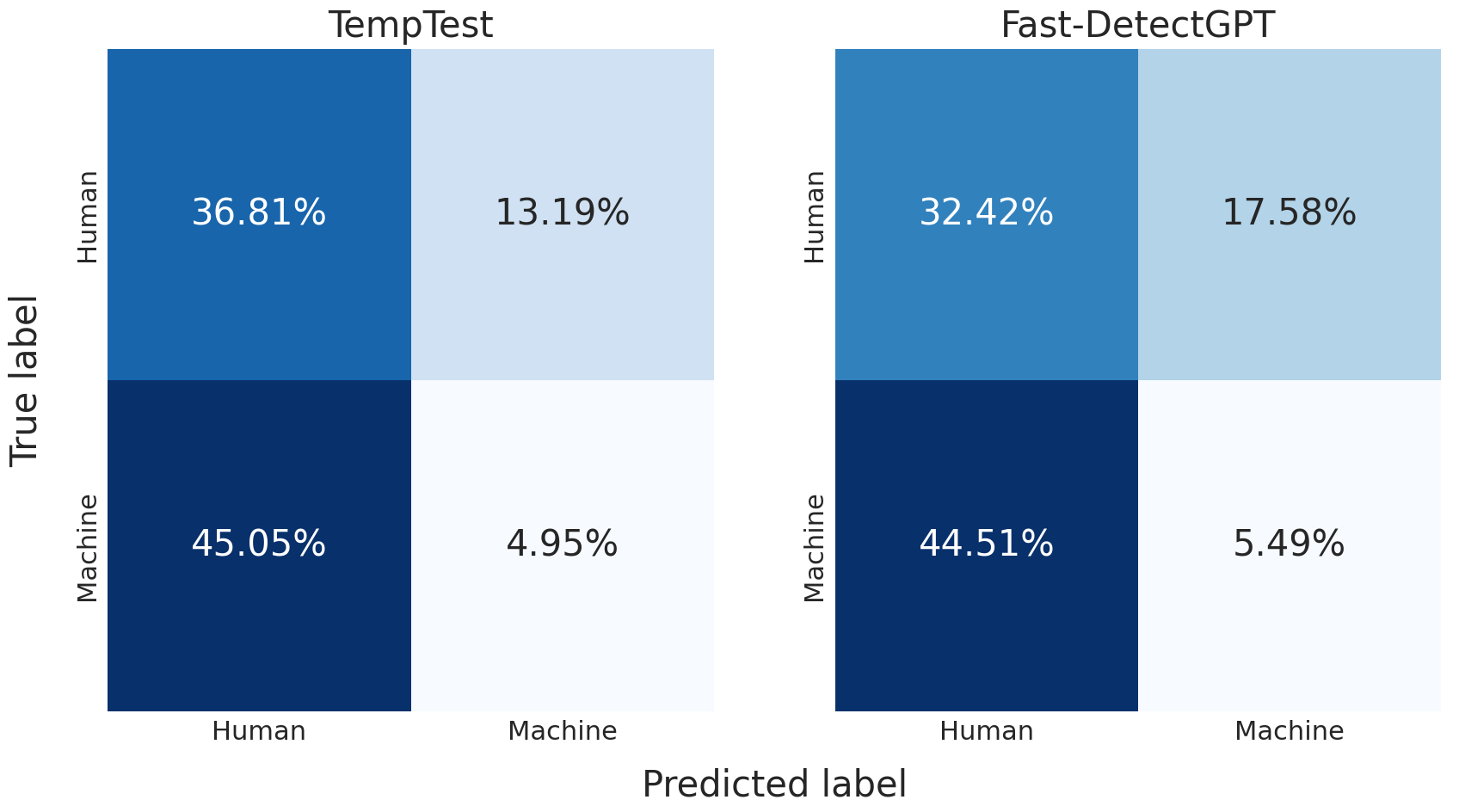}
    \vspace{.3in}
    \caption{TempTest Reduces Bias Against Non-native Speakers. Confusion matrices for performance of TempTest and Fast-DetectGPT on a dataset of TOEFL essays by non-native speakers and machine-polished adaptations. Classification was done over $50$ tokens and at an EER threshold tuned via an independent dataset (Writing). The model used for scoring and tuning the EER was GPT-Neo 2.7b, the empirically favored model of Fast-DetectGPT. TempTest has improved overall performance, and notably, the number of human essays flagged as machine is reduced. This preliminary result suggests TempTest may be effective at reducing some sources of bias in machine-generated text detection. Both methods perform poorly on the machine-polished text, indicating further research should be done to combat this form of paraphrasing attack.}
    \label{fig:non-native}
\end{figure}

Figure \ref{fig:non-native} gives the full confusion matrix for the experiment described in Section \ref{sec:non-native}.

\subsection{Nucleus Sampling}\label{app:topp}
There is a natural analogue of our top-k approach to detect text generated using nucleus (top-p) sampling \citep{Holtzman2020The}. Nucleus sampling is similar to top-k sampling, except that instead of restricting to the $k$ most likely tokens, it restricts to the smallest set of most likely tokens whose cumulative probability exceeds a threshold $p$. If the cumulative probability was always exactly $p$, then there would be no local normalization distortion, but since this is not true, one can follow the method of Section \ref{app:top-k} with the mass of the top-k set replaced with the mass of the top-p set. Unfortunately, the signal one is hoping to detect is much weaker here, and so we were not able to produce analogous empirical results.

\subsection{Algorithmic Complexity}\label{sec:complexity}
The dominant and most costly part of TempTest is performing a single forward pass with the scoring language model to compute the conditional probabilities. Therefore, TempTest inherits the chosen model's time and memory complexity.

\vfill

\end{document}